\documentclass[9 pt] {article}
\usepackage{spconf,amsmath,graphicx}
\usepackage{titlesec}
\usepackage{float} 

\titlespacing*{\section}{0pt}{6pt}{6pt}
\titlespacing*{\subsection}{0pt}{3pt}{3pt}

\usepackage{enumitem}
\setlist{nosep, leftmargin=14pt}

\usepackage{mwe} 

\def\x{{\mathbf x}}

\makeatletter

\renewenvironment{thebibliography}[1]{%
  \begin{oldthebibliography}{#1}%
  \setlength{\itemsep}{0pt}
  \setlength{\parskip}{0pt}
}{%
  \end{oldthebibliography}%
}
\makeatother

\title{Shortcut Learning in Glomerular AI: Adversarial Penalties Hurt, Entropy Helps
}
\name{%
\shortstack[c]{%
Mohammad Daouk$^{1}$, Jan Ulrich Becker$^{2}$, Neeraja Kambham$^{3}$,\\
Anthony Chang$^{4}$, Hien Nguyen$^{1,*}$, Chandra Mohan$^{1,*}$%
}}

\address{%
$^{1}$University of Houston, Houston, TX, USA\\
$^{2}$University Hospital Cologne, Cologne, Germany\\
$^{3}$Stanford University, Stanford, CA, USA\\
$^{4}$The University of Chicago, Chicago, IL, USA%
}

\begin{document}
%
\maketitle
\begingroup
\renewcommand\thefootnote{*}
\footnotetext{These authors jointly supervised this work.}
\endgroup
\begin{abstract}
Stain variability is a pervasive source of distribution shift and potential shortcut learning in renal pathology AI. We ask whether lupus nephritis glomerular lesion classifiers exploit stain as a shortcut, and how to mitigate such bias without stain or site labels. We curate a multi-center, multi-stain dataset of 9{,}674 glomerular patches (224$\times$224) from 365 WSIs across three centers and four stains (PAS, H\&E, Jones, Trichrome), labeled as proliferative vs.\ non-proliferative. We evaluate Bayesian CNN and ViT backbones with Monte Carlo dropout in three settings: (1) stain-only classification; (2) a dual-head model jointly predicting lesion and stain with supervised stain loss; and (3) a dual-head model with label-free stain regularization via entropy maximization on the stain head. In (1), stain identity is trivially learnable, confirming a strong candidate shortcut. In (2), varying the strength and sign of stain supervision strongly modulates stain performance but leaves lesion metrics essentially unchanged, indicating no measurable stain-driven shortcut learning on this multi-stain, multi-center dataset, while overly adversarial stain penalties inflate predictive uncertainty. In (3), entropy-based regularization holds stain predictions near chance without degrading lesion accuracy or calibration. Overall, a carefully curated multi-stain dataset can be inherently robust to stain shortcuts, and a Bayesian dual-head architecture with label-free entropy regularization offers a simple, deployment-friendly safeguard against potential stain-related drift in glomerular AI.
\end{abstract}


\section{Introduction}
\label{sec:intro}

Histopathological evaluation of kidney biopsies is essential yet labor-intensive and variable. Deep learning has enabled automated detection and classification in whole-slide images (WSIs); in renal pathology, models have matched or exceeded expert performance in selected tasks~\cite{ref1}. For glomeruli, CNNs achieve strong detection across several stains and promising lesion classification performance~\cite{ref2,ref3,ref4,ref6,ref7}.

However, stain and site variation (processing, protocols, scanners) produce color and texture shifts that impair generalization; stain normalization and augmentation only partly resolve this and can be task-dependent~\cite{ref8,ref9,ref10}. Even after color handling, residual site-specific signatures can remain and bias predictions~\cite{ref11}. Such signals may drive \emph{shortcut learning}, where models rely on convenient but spurious cues (e.g., stain or site) rather than underlying pathology, threatening fairness and robustness.

In lupus nephritis, glomerular lesions are graded across multiple stains (PAS, H\&E, Jones, Trichrome). Glomerular AI models trained on such data could, in principle, exploit stain identity as a shortcut when predicting proliferative vs.\ non-proliferative lesions. It remains unclear (i) whether modern models actually rely on stain in this setting and (ii) how to mitigate such shortcuts without requiring stain/site labels or image translation pipelines at deployment.

In this work we study shortcut learning from stain in lupus nephritis glomerular lesion classification using a large, multi-center, multi-stain dataset and a Bayesian dual-head architecture that explicitly probes and regularizes stain information. We adapt shortcut-testing ideas to treat stain as a putative confounder and propose a label-free stain regularization based on entropy maximization.

Our contributions are threefold:
\begin{itemize}
    \item We perform, to our knowledge, the first systematic study of stain-based shortcut learning for lupus nephritis glomerular lesion classification on a multi-center, multi-stain dataset of 9{,}674 glomerular patches labeled as proliferative vs.\ non-proliferative.
    \item We introduce a Bayesian dual-head framework (lesion + stain) that uses loss re-weighting to test the coupling between stain predictability and lesion performance, while monitoring predictive uncertainty via Monte Carlo dropout.
    \item We propose a practical, label-free stain regularization based on entropy maximization on a stain head, enforcing stain invariance without stain or site labels and without image translation, and show that it preserves lesion accuracy and calibrated uncertainty.
\end{itemize}

\section{Related Work}
\label{sec:related}

Glomerular lesion classification has progressed from classical KNN approaches ($\sim$88\% accuracy for proliferative lesions) to CNN-based systems reaching $\sim$90--93\% for non-sclerosed vs.\ sclerosed glomeruli and improving clinical workflows~\cite{ref1,ref4,ref6,ref7}. These works primarily optimize in-domain accuracy, with limited focus on robustness to stain or site variation. Cross-center color shifts are known to harm generalization; color augmentation and normalization help but are task- and dataset-dependent, while RandStainNA / RandStainNA++ unify normalization and augmentation to encourage stain-agnostic models~\cite{ref8,ref9,ref19,ref20}. Nevertheless, residual site signatures can persist and act as shortcuts~\cite{ref11}, motivating methods that explicitly suppress domain cues.

Bias mitigation strategies include in-processing domain-adversarial methods such as DANN, which reduce domain predictability but require domain labels and careful tuning~\cite{ref14,ref15}; image-level stain-transfer and GAN-based augmentation, which can reduce stain sensitivity but add complexity and potential artifacts~\cite{ref17,ref18}; and augmentation-based domain generalization, where RandStainNA/++ often outperform either normalization or augmentation alone~\cite{ref19,ref20}. IRM-style approaches such as ReConfirm suppress shortcut cues without confounders at test time~\cite{ref21}. In contrast, we focus specifically on stain as a putative shortcut for glomerular lesion classification and combine explicit shortcut testing via dual-head loss re-weighting with a label-free entropy objective on a stain head, requiring neither stain/site labels nor image translation.

\section{Methodology}
\label{sec:methods}

\subsection{Problem formulation}

We consider patch-level classification of lupus nephritis glomeruli into proliferative vs.\ non-proliferative lesions. Each patch $\x$ is associated with a lesion label $y$ and a stain label $s \in \{\text{PAS, H\&E, Jones, Trichrome}\}$. Our aims are to (i) test whether deep models exploit $s$ as a shortcut when predicting $y$, and (ii) learn stain-agnostic representations that preserve lesion performance, without requiring stain labels at deployment.

\subsection{Bayesian backbones and uncertainty}

We treat a family of convolutional and transformer backbones (ResNet, DenseNet, EfficientNet, RegNetY, ResNeXt, ViT) as Bayesian neural networks via Monte Carlo (MC) dropout at inference ($T=50$). Predictive uncertainty is estimated as the variance of the $T$ stochastic forward passes. This simple Bayesian approximation allows us to track how stain-related interventions affect not only accuracy but also uncertainty, which we use as an early warning signal for destabilized representations.

\subsection{Dual-head architecture}

We build a shared feature extractor $f_\theta$ followed by two task-specific heads: a lesion head $h_\ell$ and a stain head $h_s$. Given a patch $\x$, the shared trunk produces features $z = f_\theta(\x)$; the lesion head outputs $p_\ell(y \mid z)$ and the stain head outputs $p_s(s \mid z)$. We compare a single-head model that predicts stain only (Experiment~1) and a dual-head model that jointly predicts lesion and stain (Experiments~2--3). The total loss for the dual-head model is
\begin{equation}
    \mathcal{L} = \mu_{1} \, \mathcal{L}_{\text{lesion}} + \mu_{2} \, \mathcal{L}_{\text{stain}},
\end{equation}
where $\mathcal{L}_{\text{lesion}}$ is cross-entropy on lesion labels, and $\mathcal{L}_{\text{stain}}$ differs by experiment.

\subsection{Shortcut testing via loss re-weighting}

To test whether stain acts as a shortcut, we follow the shortcut-testing paradigm of Brown \emph{et al.}~\cite{ref22} and treat $\mu_2$ as a knob that makes stain easier or harder to learn. In Experiment~2, we set $\mathcal{L}_{\text{stain}}$ to supervised cross-entropy on stain labels and sweep $\mu_2$ over positive, zero, and negative values. Positive $\mu_2$ rewards accurate stain prediction, whereas negative $\mu_2$ penalizes it. If lesion performance improves when stain is emphasized and degrades when stain is suppressed, this indicates reliance on stain as a shortcut; flat lesion performance across $\mu_2$ suggests robustness to stain cues.

\subsection{Label-free stain entropy regularization}

In Experiment~3, we replace supervised stain cross-entropy with Reverse Cross-Entropy (RCE), implemented as entropy maximization on the stain head predictions:
\begin{equation}
    H(p^{\text{stain}}) = - \sum_{k=1}^{K} p^{\text{stain}}_{k} \, \log p^{\text{stain}}_{k}.
\end{equation}
We use $\mu_2 \le 0$ so that minimizing $\mathcal{L}$ encourages high-entropy (near-uniform) stain predictions, effectively pushing the representation to become stain-invariant. Crucially, this objective does not require stain labels and can be applied at deployment time when stain or site metadata may be unavailable.

\subsection{Training and evaluation protocol}

Data are split at the WSI level: 85\% development (train/validation via stratified 5-fold cross-validation) and 15\% held-out test. Each backbone produces five cross validation (CV) variants. We optimize with Adam (batch 32), ReduceLROnPlateau scheduler, and early stopping (patience 7, max 50 epochs) implemented on validation loss, using mixed precision for efficiency on an NVIDIA Quadro RTX 8000. Metrics include accuracy, precision, recall, F1, AUC, and Bayesian predictive uncertainty.

\section{Experiments and Results}
\label{sec:experiments}

\subsection{Dataset}
\label{subsec:data}

We curated 9{,}674 glomerular image patches (224$\times$224 px) from 365 WSIs across three centers (University Hospital of Cologne, Stanford University, University of Chicago). Patches span four stains; PAS (2{,}412; 24.9\%), H\&E (2{,}252; 23.3\%), Jones (2{,}270; 23.5\%), and Trichrome (2{,}740; 28.3\%), and are labeled as proliferative (1{,}907; 19.7\%) or non-proliferative (7{,}767; 80.3\%). This multi-center, multi-stain dataset provides a realistic testbed for stain-related shortcut learning.

\subsection{Experiment 1: Can the model recognize stain type?}

We first train a single-head Bayesian classifier to predict stain (PAS/H\&E/Jones/Trichrome) from glomerular patches. Across backbones, performance is near-perfect (accuracy/AUC/-precision/recall/F1 $\approx 1.0$) with very low predictive uncertainty ($\sim 0.001$). This confirms that stain identity is trivially learnable from color/texture cues and thus represents a plausible shortcut for downstream tasks.

\subsection{Experiment 2: Does stain supervision induce shortcuts for lesion?}

We next train a dual-head model with supervised lesion cross-entropy and supervised stain cross-entropy, sweeping $\mu_2$ over positive, zero, and negative values as in Section~\ref{sec:methods}. Positive $\mu_2$ increases stain accuracy and AUC; negative $\mu_2$ drives stain performance below chance, consistent with learning an inverted label signal. At $\mu_2 = 0$, stain AUC is $\approx 0.5$ with accuracy/precision/recall $\approx 0.25$ (chance level) and F1 $\approx 0.20$ (slightly under chance for four classes).

Across models and folds, lesion accuracy, AUC, precision, recall, and F1 show no significant changes as $\mu_2$ varies, indicating no measurable shortcut learning from stain to lesion classification on this multi-stain, multi-center dataset. However, lesion predictive uncertainty increases as $\mu_2$ becomes more negative (from $\sim$0.06 at $\mu_2 = 0$ to $\sim$0.5 when stain performance is pushed far below chance) and is negatively correlated with lesion accuracy. Penalizing stain prediction thus destabilizes lesion feature learning, raising uncertainty without improving lesion metrics.

\begin{figure}[t]
\centering
\includegraphics[width=1.1\linewidth]{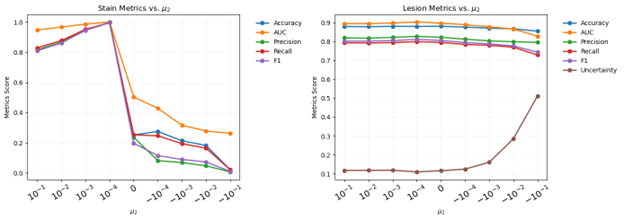}
\caption{\textbf{Experiment 2.} Positive $\mu_{2}$ boosts stain metrics; strongly negative $\mu_{2}$ (adversarial CE) harms stain metrics and raises lesion predictive uncertainty.}
\label{fig:exp2}
\end{figure}

\subsection{Experiment 3: Label-free mitigation of stain features}

Finally, we retain the dual-head architecture but replace supervised stain loss with entropy maximization (label-free RCE), sweeping $\mu_2 \le 0$ in a moderate range. Stain F1 is pulled back to chance ($\approx 0.25$), and AUC/accuracy/precision/-recall remain at chance; unlike Experiment~2, stain performance does not collapse below chance. Lesion metrics remain stable and indistinguishable from the $\mu_2 = 0$ baseline, and lesion predictive uncertainty stays low ($\sim 0.06$), avoiding the inflation seen with negative-weighted supervised stain loss.

\begin{figure}[t]
\centering
\includegraphics[width=0.85\linewidth]{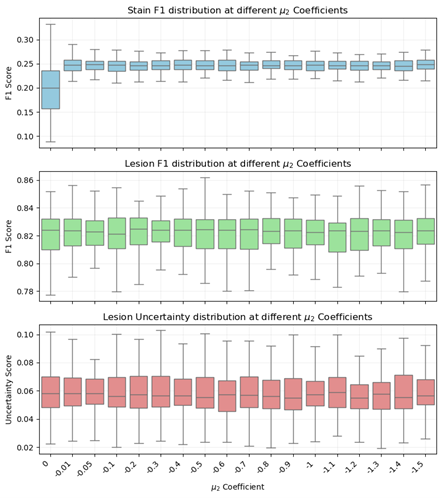}
\caption{\textbf{Experiment 3.} Label-free RCE drives stain predictions to chance without altering lesion performance or uncertainty.}
\label{fig:exp3}
\end{figure}

These results indicate that label-free entropy regularization can remove stain information without adversarial over-correction and without harming lesion performance or confidence.

\section{Discussion and Conclusion}
\label{sec:discussion}

\textbf{Stain is easy; shortcuts are not inevitable.} Models can perfectly recognize stain (Experiment~1), yet lesion classification on our multi-stain, multi-center dataset does not depend on stain cues (Experiment~2). This suggests that dataset diversity itself discourages stain-based shortcuts for this task.

\textbf{Adversarial supervision can backfire.} Driving a supervised stain head with strongly negative $\mu_2$ forces stain predictions below chance and inflates lesion uncertainty, while offering no improvement in lesion metrics. This behavior is consistent with known pathologies of adversarial training with explicit labels.

\textbf{Label-free entropy is a safer safeguard.} Replacing supervised stain cross-entropy with entropy maximization (Experiment~3) produces stain-agnostic representations that hold stain performance at chance, keep lesion metrics unchanged, and preserve calibrated uncertainty. This achieves the goal of mitigating potential shortcut signals without access to stain labels and without degrading lesion accuracy.

\textbf{Bayesian uncertainty as an early warning.} Uncertainty rises when adversarial pressure destabilizes representations (Experiment~2) and stays low when the mitigation is benign (Experiment~3), suggesting that predictive variance is a useful diagnostic for monitoring representational drift under domain-invariance interventions.

\textbf{Practical takeaway.} For lupus nephritis glomerular lesion classification, a carefully curated multi-stain, multi-center dataset can already be robust against stain-driven shortcuts. When additional safeguards are desired (e.g., for deployment under future stain/site shifts), a Bayesian dual-head model with label-free stain entropy regularization offers a simple, unsupervised, and deployment-friendly strategy to mitigate potential stain-related drift while preserving accuracy and confidence.

\section{Compliance with Ethical Standards}

This study was performed in line with the principles of the Declaration of Helsinki. The retrospective use of de-identified human subject data was approved by the institutional review boards (IRBs) of the University of Houston, University Hospital Cologne, Stanford University, and the University of Chicago. All data were anonymized prior to analysis.

\section{Acknowledgments}
\label{sec:acknowledgments}

This work was supported by NIH R01DK134055.

Dr. Mohan has consultancy or sponsored research agreements or equity with Boehringer-Ingelheim, Progentec Diagnostics, and Voyager Therapeutics. Dr. Mohan is on the Medical Scientific Advisory Council of the Lupus Foundation of America.
Dr. Mohan’s research is supported by NIH RO1 AR074096 and DK134055.

{\fontsize{9}{10}\selectfont
\bibliographystyle{IEEEbib}
\bibliography{main}
}

\end{document}